\documentclass[letterpaper]{article} % DO NOT CHANGE THIS
\PassOptionsToPackage{table,xcdraw}{xcolor}
\usepackage[preprint]{arxiv}
\usepackage[hyphens]{url}  % DO NOT CHANGE THIS
\usepackage{graphicx} % DO NOT CHANGE THIS
\usepackage{natbib}  % DO NOT CHANGE THIS AND DO NOT ADD ANY OPTIONS TO IT
\usepackage{caption} % DO NOT CHANGE THIS AND DO NOT ADD ANY OPTIONS TO IT
\usepackage{algorithm}
\usepackage{algorithmic}

\usepackage{multirow}
\usepackage{xcolor}

\usepackage{amsmath}
\usepackage{amssymb}
\usepackage{mathtools}
\usepackage{amsthm}

\usepackage{newfloat}
\usepackage{listings}
\DeclareCaptionStyle{ruled}{labelfont=normalfont,labelsep=colon,strut=off} % DO NOT CHANGE THIS
\floatstyle{ruled}
\newfloat{listing}{tb}{lst}{}
\floatname{listing}{Listing}

\usepackage{booktabs}

\title{SemMask: Semantic-Prioritized Progressive Masking for \\ Diffusion Language Models}
\author{
    Qi Chen\textsuperscript{\rm 2,3}\equalcontrib, Zifeng Cheng\textsuperscript{\rm 1}\equalcontrib, Zhiwei Jiang\textsuperscript{\rm 1}\equalcontrib\corresponding, Qing Gu\textsuperscript{\rm 1}\\
}
\affiliations{
    \textsuperscript{\rm 1}State Key Laboratory for Novel Software Technology, Nanjing University, China\\
\textsuperscript{\rm 2}University of Chinese Academy of Sciences, China\\
\textsuperscript{\rm 3}Shenyang Institute of Computing Technology, Chinese Academy of Sciences, China \\
    chenqi231@mails.ucas.ac.cn, chengzf@smail.nju.edu.cn, \{jzw,guq\}@nju.edu.cn
}

\begin{document}

\maketitle

%% --------- Abstract --------- %%
\begin{abstract}
Diffusion Language Models (DLMs) are designed to enable globally parallel generation. However, in practice, they often exhibit pseudo-autoregressive behavior and a function-word-first bias, which undermines their performance on tasks such as text writing and code generation that benefit from global planning and parallel decoding. To address this, we propose SemMask, a novel fine-tuning paradigm that leverages a semantic-prioritized progressive masking strategy to encourage the model to decode high-information semantic tokens early and globally, thereby enhancing its global planning and parallel generation capabilities. Specifically, we first use the gradients of the original DLM with respect to each token to identify semantic tokens. During masking, these semantic tokens are preserved with priority, establishing a progressive generation order in which semantic tokens are decoded prior to function tokens. Meanwhile, we adjust the loss weight for semantic tokens to further reinforce their early prediction. Experimental results show that SemMask achieves quality comparable to strong autoregressive baselines on both text writing and code generation tasks, while providing superior semantic coherence. Notably, SemMask achieves higher accuracy with substantially fewer decoding steps, indicating that the early establishment of semantic structure improves the efficiency--accuracy trade-off.
\end{abstract}
%% --------- END --------- %%


%% --------- Introduction --------- %%
\section{Introduction}

\begin{figure*}[h]
    \centering
    \includegraphics[width=\linewidth]{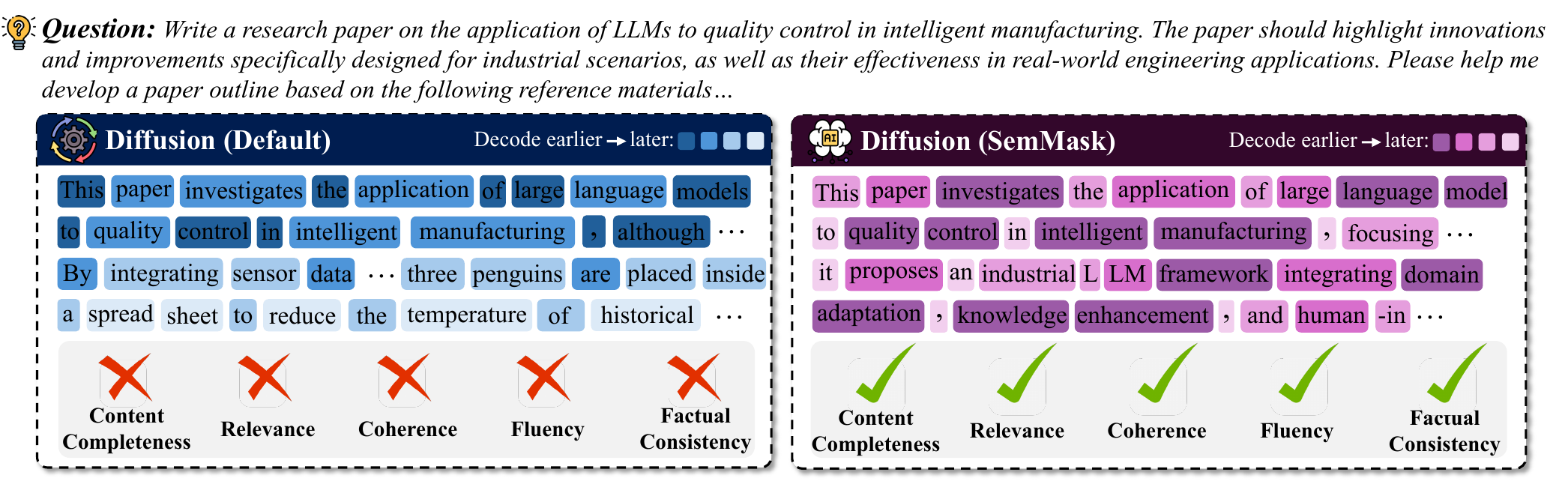}
    % \caption{(Left) Standard DLMs exhibit an pseudo-autoregressive behavior, where the generation trajectory collapses into a linear diagonal, coupling generation order with token position. (Right) \textbf{SemMask} establishes a global generation paradigm: it explicitly prioritizes semantic tokens (blue crosses) in the early generation steps to construct the global structure, before filling in function tokens (green circles) to complete local details.}
    \caption{Default diffusion decoding exhibits pseudo-autoregressive behavior, resulting in incomplete and structurally disorganized content, whereas SemMask prioritizes semantically important tokens to establish a global semantic scaffold and generate more coherent and complete text.}
    % \vspace{-2mm} 参考liu 正好论文
    \label{fig:intro}
\end{figure*}

In recent years, Diffusion Language Models (DLMs) have emerged as a promising alternative to traditional autoregressive (AR) models, enabling parallel and efficient text generation~\cite{austin2021structured,li2022diffusion,gong2025scaling,zhang2025diffusion}.
Unlike AR models that generate tokens sequentially from left to right, DLMs iteratively generate all tokens in parallel, enabling more global generation planning~\cite{ye2025dream,nie2026large}.

\begin{figure}[t]
    \centering
    \includegraphics[width=\linewidth]{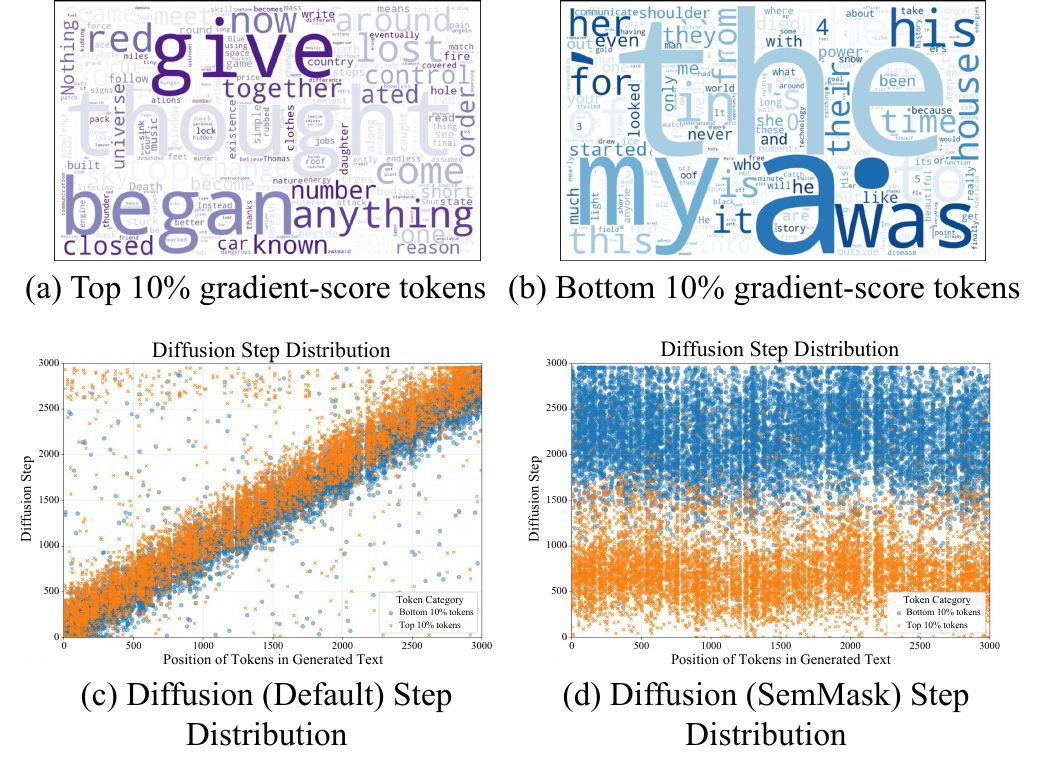}
    % \caption{High-gradient words (nouns and verbs) and low-gradient words (conjunctions, prepositions, articles, and punctuation) calculated on the WritingPrompts dataset.}
    \caption{Analysis of the function-word-first bias on HANNA using LLaDA-8B-Instruct, with both the generation length and the number of sampling steps set to 3,000.}
    % \vspace{-2mm}
    \label{fig:token}
\end{figure}

However, a significant gap remains between the architectural potential of DLMs and their practical decoding behavior~\cite{deschenaux2024promises,xu2025energy,feng2026theoretical}. 
While DLMs are designed to exploit bidirectional context and decode tokens in parallel, they exhibit two systematic decoding phenomena in practice: pseudo-autoregressive behavior and a function-word-first bias.
As shown in Figure~\ref{fig:intro}, under the default pure diffusion sampling strategy, DLMs exhibit a pseudo-autoregressive decoding behavior. 
This behavior compromises content completeness, resulting in fragmented and structurally disorganized generation.
% To illustrate this phenomenon, we conduct controlled experiments with LLaDA-8B-Instruct on the HANNA text writing task, as shown in Figure~\ref{fig:token}. % under pure diffusion sampling 写在caption 图2
To empirically examine the function-word-first bias, we conduct a series of controlled experiments, with the results shown in Figure~\ref{fig:token}.
As shown in Figure~\ref{fig:token}(c), the bottom 10\% gradient-score tokens, shown in blue, are consistently decoded at earlier diffusion steps than the top 10\% tokens, shown in orange. 
Figures~\ref{fig:token}(a) and (b) show that the top 10\% gradient-score tokens are primarily semantic tokens, such as nouns and verbs, whereas the bottom 10\% are mostly function tokens, including pronouns, prepositions, and articles. 
These observations indicate a clear decoding bias toward decoding low-information function tokens earlier while delaying high-information semantic tokens.

These phenomena arise from the confidence-guided remasking strategy employed during inference, which encourages DLMs to decode higher-confidence tokens while re-masking lower-confidence ones~\cite{nie2026large,huang2025don}.
For DLMs trained with a random masking strategy, these higher-confidence tokens are often either early tokens in the sequence or function words, leading to a pseudo-autoregressive, function-word-first generation order.
This generation order undermines the inherent global planning advantage of DLMs, resulting in suboptimal performance on tasks that rely on such global planning, such as text writing and code generation.

To address these issues, we propose SemMask, a novel fine-tuning paradigm designed to encourage the model to decode high-information semantic tokens early and globally, thereby activating its global planning and parallel generation capabilities. 
As illustrated in Figure~\ref{fig:intro} and Figure~\ref{fig:token}(d), our approach dissolves the pseudo-autoregressive behavior and function-word-first bias observed in the original model.
To achieve this goal, we focus on solving two core problems: (1) how to identify the high-information semantic tokens in a text; and (2) how to guide the model to perform semantic-prioritized generation.
For the first problem, we design a Gradient-Based Semantic Token Scoring strategy, which utilizes the gradients of the original DLM with respect to each token to identify semantic tokens.
For the second problem, we design a Semantic Anchor Preservation Masking strategy for mask generation and a Semantic-Prioritized Loss Re-Weighting strategy for objective adjustment.
During training, the identified semantic tokens are preserved with priority, establishing a progressive generation order in which semantic tokens are decoded prior to function tokens. 
During objective adjustment, the weights of semantic tokens in the final loss are also adjusted to further reinforce their early prediction.

To verify the effectiveness of our proposed SemMask, we conduct extensive experiments on both text writing and code generation benchmarks. 
The experimental results show that SemMask achieves performance comparable to strong autoregressive baselines on both tasks, while providing superior semantic coherence and inference efficiency. 
% 加上实验再重写
Notably, SemMask maintains strong performance with substantially fewer decoding steps, confirming that the early establishment of a semantic scaffold restores genuine parallel decoding capability.

Our contributions are summarized as follows:
\begin{itemize}
    \item We propose a novel training paradigm that strengthens the DLM's ability to predict semantic tokens early in the generation process, thereby activating its global planning and parallel generation capabilities.
    \item We introduce \textbf{Gradient-Based Semantic Token Scoring}, where loss gradients are used to quantify token importance, enabling the model to prioritize hard-to-learn semantic tokens over function tokens.
    \item We introduce the \textbf{Semantic Anchor Preservation Masking}, a sparsity-constrained masking strategy that preserves the spatial integrity of semantic tokens during high-noise states. This mechanism enforces a global generation approach by ensuring that the global structure of the sequence is reconstructed before local syntactic details are filled in.
\end{itemize}
%% --------- END --------- %%

\begin{figure*}[h]
    \centering
    \includegraphics[width=\linewidth]{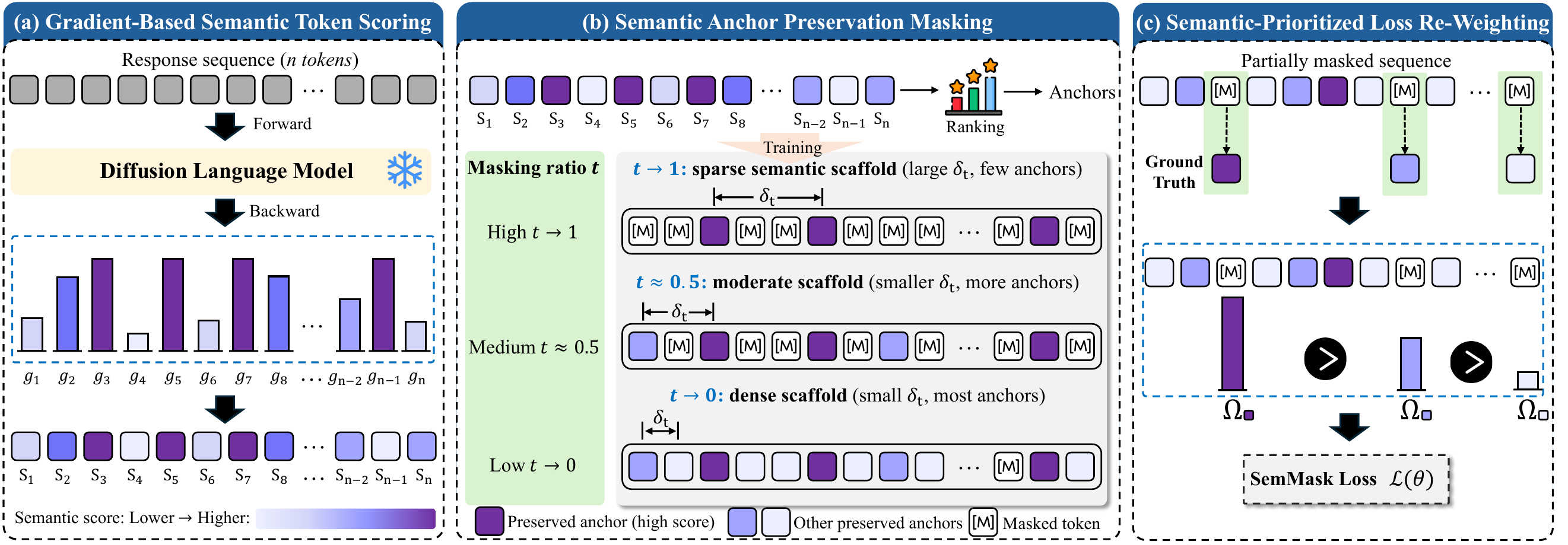}
    \caption{Overview of the SemMask framework. (a) \textbf{Gradient-Based Semantic Token Scoring} estimates the semantic importance of each token. (b) \textbf{Semantic Anchor Preservation Masking} ranks tokens by their importance scores and progressively preserves high-scoring tokens as semantic anchors. (c) \textbf{Semantic-Prioritized Loss Re-Weighting} assigns larger optimization weights to more important tokens to encourage their early reconstruction.}
    \label{fig:method}
    % \vspace{-1mm}
\end{figure*}

%% --------- Related Work ---------%%
\section{Related Work}

\subsection{Diffusion Language Models}
Diffusion language models (DLMs) have recently emerged as a promising alternative to traditional autoregressive models for text generation.
While early approaches~\cite{li2022diffusion,dieleman2022continuous,gao2024empowering} focused on exploring continuous embedding spaces, the field has increasingly shifted towards Masked Diffusion Models (MDMs), which directly model language in the categorical state space by predicting masked tokens through iterative denoising, enabling fast and parallel decoding~\cite{austin2021structured,sahoo2024simple,hersche2025soft}. 
Building upon this discrete formulation, recent advancements have demonstrated that DLMs can rival autoregressive baselines at scale. 
A notable example is LLaDA, trained from scratch, which achieves competitive performance against strong baselines such as LLaMA 3‑8B~\cite{nie2026large}. 
Similarly, Dream presents a family of scalable diffusion models that match state-of‑the‑art autoregressive performance across general, mathematical, and coding tasks~\cite{ye2025dream}. 
Beyond general-purpose generation, recent research has expanded DLMs into various multimodal domains~\cite{yang2026mmada,li2026lavida,you2026llada} and specialized reasoning tasks, including code generation and mathematical inference~\cite{xie2025dream,huang2026reinforcing}. 

Despite these rapid advances, the prevailing paradigm in discrete DLMs remains uniform corruption, where tokens are masked uniformly at random. This agnostic masking policy treats all tokens identically, overlooking the hierarchical information density of natural language~\cite{chao2026beyond,yu2025discrete}, which often causes models to struggle with capturing global semantic dependencies in the early stages of generation. Furthermore, this homogeneous nature fails to provide explicit guidance on the optimal generation order, rendering naive decoding strategies highly sensitive to local context~\cite{hong2025improving}, thereby limiting the practical realization of true non‑autoregressive modeling.

\subsection{Non-Autoregressive Generation and Token Ordering}
Non-Autoregressive generation aims to circumvent the high inference latency inherent in Autoregressive models by decoding multiple tokens in parallel. 
Seminal works like Insertion Transformers~\cite{stern2019insertion} and FELIX~\cite{mallinson2020felix} explored permutation and dynamic insertion to break rigid sequential dependencies. 
In the context of MDMs, the noise schedule effectively serves as an implicit generation order. 
However, despite the theoretical capability for global parallel planning, position bias causes tokens adjacent to the left context to exhibit lower entropy and higher confidence, leading to a left-to-right convergence pattern~\cite{gwak2025reward,chen2025dparallel}. 
Consequently, early prediction errors propagate through subsequent tokens, often resulting in semantic drift and reduced coherence in longer sequences, akin to the error accumulation commonly observed in autoregressive decoding~\cite{madjidi2023survey,deschenaux2024promises,kowsher2025predicting}.

To restore true parallel reasoning, recent research has focused on optimizing token ordering through hybrid architectures, trajectory alignment, and adaptive inference. 
Structurally, methods like Block Diffusion~\cite{arriola2025block}, Seed Diffusion~\cite{song2025seed}, and ReFusion~\cite{li2025refusion} introduce semi-autoregressive block or slot-based mechanisms to enforce coarse-grained consistency. 
Beyond architectural constraints, several works optimize the generation trajectory directly during training: dParallel~\cite{chen2025dparallel} employs distillation to enforce deterministic convergence, while DiffuCoder~\cite{gong2025diffucoder} and TraceRL~\cite{wang2025revolutionizing} align training objectives with inference paths via reinforcement learning. 
Additionally, approaches like Prophet~\cite{li2025diffusion} and KLASS~\cite{kim2026klass} exploit confidence gaps and stability metrics for early exit, whereas Dilated Scheduling~\cite{luxembourg2025plan} and adaptive sub-problem selection~\cite{kim2025train} optimize unmasking sequences to mitigate entropy accumulation.
However, these approaches largely remain within the paradigm of parallelization of independent sampling rather than parallelization of semantic association. 
Therefore, our approach internalizes semantic priority decisions within the model, guiding it to first establish a global semantic scaffold by predicting semantic tokens with high information content across the sequence and then progressively fill in the remaining local function tokens.

%% --------- END --------- %%

%% ----- Methodology ----- %%
\section{Methodology}
Our approach, SemMask, consists of three stages: (a) Gradient-Based Semantic Token Scoring estimates the semantic importance of each token, (b) Semantic Anchor Preservation Masking ranks tokens by their importance scores and progressively preserves them according to these scores, and (c) Semantic-Prioritized Loss Re-Weighting assigns greater optimization weights to tokens with higher importance scores to encourage their early reconstruction. 
Together, these components guide the model to first establish a global semantic scaffold and then progressively complete the remaining tokens, as illustrated in Figure~\ref{fig:method}.

\subsection{Gradient-Based Semantic Token Scoring}
Default masking strategies fail to account for the varying importance of tokens in text.
To address this limitation, we assume that the semantic importance of a token is reflected in the sensitivity of the model’s predictions to changes in that token.
Accordingly, we use the magnitude of its gradient to quantify token importance.

Given a text sequence $\boldsymbol{x}=(x_1,\ldots,x_n)$, we use the frozen parameters of the DLM to compute the gradient $g_i$ for each token $x_i$.
To ensure a consistent scale for token importance scores across different text sequences, we normalize the raw scores to obtain the final score $S_i$:
\begin{equation}
\label{eq:semantic_score}
S_i =
\frac{
g_i - \min(\mathcal{G})
}{
\max(\mathcal{G}) - \min(\mathcal{G}) + \epsilon
}
\end{equation}
where $\mathcal{G}={g_i}_{i=1}^{n}$ denotes the set of gradient values for all tokens in the sequence $\boldsymbol{x}$, $\min(\mathcal{G})$ and $\max(\mathcal{G})$ denote the minimum and maximum values in $\mathcal{G}$, respectively, and $\epsilon$ is a small constant introduced for numerical stability.

\begin{algorithm}[tb]
    \caption{Semantic Anchor Preservation Masking}
    \label{alg:SemMask}
    
    \textbf{Input}: Clean sequence $\boldsymbol{x}_0$, semantic scores $\mathcal{S}$, masking ratio $t$\\
    \textbf{Parameter}: Anchor distance $\delta$, perturbation strength $\lambda$ \\
    \textbf{Output}: Partially masked sequence $\boldsymbol{x}_t$
    
    \begin{algorithmic}[1]
        \STATE Initialize anchor set $\mathcal{A}\leftarrow\emptyset$
        and forbidden set $\mathcal{F}\leftarrow\emptyset$
        \STATE Set the target anchor count
        $K\leftarrow\left\lfloor L(1-t)\right\rfloor$
        \STATE Set the anchor distance
        $\delta_t\leftarrow\left\lfloor\delta t\right\rfloor$
        
        \FOR{$i=1$ to $L$}
            \STATE Sample $z_i\sim\operatorname{Gumbel}(0,1)$
            \STATE Compute $q_i\leftarrow S_i+\lambda z_i$
        \ENDFOR
        \STATE Let $\mathcal{O}$ be the indices sorted by $q_i$ in descending order
        \FOR{$i$ in $\mathcal{O}$}
            \IF{$|\mathcal{A}|<K$ and $i\notin\mathcal{F}$}
                \STATE $\mathcal{A}\leftarrow\mathcal{A}\cup\{i\}$
                \STATE $\mathcal{F}\leftarrow\mathcal{F}\cup
                \{j\in\{1,\ldots,L\}:|j-i|<\delta_t\}$
            \ENDIF
        \ENDFOR
        \FOR{$i=1$ to $L$}
            \STATE $x_t^i\leftarrow x_0^i$ if $i\in\mathcal{A}$, otherwise $x_t^i\leftarrow\texttt{[M]}$
        \ENDFOR
        \STATE \textbf{return} $\boldsymbol{x}_t$
    \end{algorithmic}
\end{algorithm}
% \vspace{-2mm}

\subsection{Semantic Anchor Preservation Masking}
Based on the token importance scores computed in the previous section, we use these scores to guide token masking during training, prioritizing tokens with higher scores as semantic tokens.

As shown in Algorithm~\ref{alg:SemMask}, we first rank all tokens according to their importance scores and retain the highest scoring tokens as semantic anchors in the masked sequence, while the remaining tokens stay masked. 
Given the normalized semantic scores $\mathcal{S}=(S_1,\ldots,S_n)$, we select the semantic anchors to be retained by maximizing the following objective:
\begin{equation}
\mathcal{A}^{*}
=
\underset{
\substack{
\mathcal{A}\subseteq\{1,\ldots,L\}
}
}{\arg\max}
\sum_{i\in\mathcal{A}}
\left(S_i+\lambda z_i\right),
\qquad
z_i{\sim}\operatorname{Gumbel}(0,1)
\end{equation}
where $\mathcal{A}^{*}$ is the optimal anchor set, $L$ is the sequence length, $z_i$ introduces stochasticity, and $\lambda$ controls the perturbation strength to balance semantic preference and anchor-selection diversity.

However, relying solely on importance scores for anchoring may lead to anchor clustering and insufficient contextual coverage, thereby increasing the difficulty of solving the generation task.
Therefore, during the stages of training, we impose the constraints defined in Eq.~\eqref{eq:constraints} to ensure sufficient contextual coverage.
\begin{equation}
\label{eq:constraints}
\resizebox{1.02\columnwidth}{!}{$
\left\{
\begin{array}{ll}
|\mathcal{A}|=\left\lfloor L(1-t)\right\rfloor,
& \text{Anchor Cardinality}\\[2pt]
|i-j|\geq\delta_t,\quad
\forall i,j\in\mathcal{A},\ i\neq j,
& \text{Sequence-Space Coverage}
\end{array}
\right.
$}
\end{equation}
where $i$ and $j$ denote anchor positions in the text sequence, $t \sim \mathcal{U}(0,1)$ denotes the masking ratio independently sampled for each training example and applied independently to all tokens, and $\delta_t$ specifies the minimum distance between anchors.
As $t\to1$, the spatial constraint encourages the retained tokens to form a global semantic scaffold. 
As $t\to0$, $\delta_t$ decays to zero, allowing the remaining tokens to be progressively filled into the established global semantic scaffold.

\subsection{Semantic-Prioritized Loss Re-Weighting}
We propose a re-weighted loss function that adjusts the learning signal according to semantic scores during training, encouraging the model to prioritize the prediction of semantic tokens with higher scores.

As shown in Eq.~\eqref{eq:omega_weight}, we introduce $\Omega_i$ into the loss function by mapping each token's semantic score to its corresponding importance weight.
Here, $\alpha$ controls the overall weighting magnitude, while $\beta$ determines the emphasis placed on semantic tokens.
\begin{equation}
\label{eq:omega_weight}
\Omega_i
=
1+\alpha\cdot\exp\left(\beta S_i\right)
\end{equation}
The resulting loss function is defined as:
\begin{equation}
\resizebox{1.0\columnwidth}{!}{
$ \displaystyle
\mathcal{L}(\theta) = -\mathbb{E}_{t,x_{0},x_{t}} \left[ \frac{1}{t \cdot \mathcal{Z}_\Omega} \sum_{i=1}^{L} \mathbb{I}[x_{t}^{i} = \texttt{[M]}] \cdot \Omega_i \cdot \ell^{i}(\theta) \right]
$
}
\end{equation}
where $\ell^{i}(\theta)=\log p_{\theta}(x_0^{i}\mid \texttt{prompt}, \boldsymbol{x_t})$ denotes the token-level log-likelihood, $\mathbb{I}[\cdot]$ is the mask indicator, and $\mathcal{Z}_{\Omega}=\sum_{i\in\mathrm{masked}}\Omega_i$ normalizes the weighted objective.
Accordingly, this training objective reinforces the progressive generation order induced by the masking strategy.

%% --------- END --------- %%

\begin{table*}[ht]

\centering
\renewcommand{\arraystretch}{1.1} 

\resizebox{0.92\textwidth}{!}{%
\begin{tabular}{lccccccc}
\toprule

\multirow{2}{*}{\textbf{Model}} & \multicolumn{5}{c}{\textbf{HANNA}} & \multicolumn{2}{c}{\textbf{CommonGen}} \\
\cmidrule(lr){2-6} \cmidrule(lr){7-8} 

 & \textbf{BLEU}$\uparrow$ & \textbf{BLEURT}$\uparrow$ & \textbf{ROUGE-L}$\uparrow$ & \textbf{BERT-Score}$\uparrow$ & \textbf{GPT-5.1-Score}$\uparrow$ & \textbf{GOD}$\uparrow$ & \textbf{Perplexity}$\downarrow$ \\
\midrule

% --- Autoregressive Models ---
\multicolumn{8}{c}{\textit{Autoregressive Models}} \\
\midrule
Llama-3-8B-Instruct & 0.3109 & 0.7208 & 0.3377 & \textbf{0.8051} & 3.0833 & -- & 24.5731 \\
Qwen2.5-7B-Instruct & 0.3893 & 0.7129 & 0.3153 & \underline{0.8030} & \underline{3.1042} & -- & 22.9951 \\
\midrule

% --- Diffusion Language Models ---
\multicolumn{8}{c}{\textit{Diffusion Language Models}} \\
\midrule
LLaDA-8B-Base     & 0.3374 & 0.6936 & 0.2101 & 0.7541 & 1.9583 & 1.5624 & 49.1982 \\
LLaDA-8B-Instruct & 0.3883 & 0.7014 & 0.2125 & 0.6932 & 2.8646 & 4.3430 & 33.2763 \\
LLaDA-8B-SFT     & 0.3352 & 0.7218 & 0.3053 & 0.7592 & 2.7708 & 4.8210 & 30.4352 \\
\rowcolor[HTML]{E8E8E8} 
LLaDA-8B-SemMask (Ours) & 0.3921 & 0.7301 & 0.3101 & 0.7746 & 3.0104 & \underline{6.3187} & \underline{22.5286} \\
\midrule
Dream-7B-Base     & 0.3640 & \underline{0.7472} & 0.3215 & 0.7664 & 2.0104 & 1.6704 & 40.9942 \\
Dream-7B-Instruct & 0.4030 & 0.6828 & 0.3762 & 0.7415 & 2.9167 & 3.8861 & 32.6030 \\
Dream-7B-SFT     & \underline{0.4086} & 0.7364 & \underline{0.3930} & 0.7526 & 3.0208 & 3.9004 & 29.1478 \\
\rowcolor[HTML]{E8E8E8} 
Dream-7B-SemMask (Ours) & \textbf{0.4613} & \textbf{0.8254} & \textbf{0.4181} & 0.7762 & \textbf{3.1250} & \textbf{7.4544} & \textbf{21.6630} \\

\bottomrule
\end{tabular}%
}
\caption{Main results of text writing evaluation on HANNA and CommonGen. The best results in each category are highlighted in \textbf{bold}, and the second-best results are marked with \underline{underlines}.}
\label{tab:writing_gen}

\end{table*}

\begin{table*}[ht]

\centering
\renewcommand{\arraystretch}{1.1} 

\resizebox{0.9\textwidth}{!}{%
\begin{tabular}{lcccccc}
\toprule

\multirow{2}{*}{\textbf{Model}} & \multicolumn{3}{c}{\textbf{HumanEval}} & \multicolumn{3}{c}{\textbf{MBPP}} \\
\cmidrule(lr){2-4} \cmidrule(lr){5-7} 

 & \textbf{Accuracy}$\uparrow$ & \textbf{CodeBLEU}$\uparrow$ & \textbf{BLEURT}$\uparrow$ & \textbf{Accuracy}$\uparrow$ & \textbf{CodeBLEU}$\uparrow$ & \textbf{BLEURT}$\uparrow$ \\
\midrule

% --- Autoregressive Models ---
\multicolumn{7}{c}{\textit{Autoregressive Models}} \\
\midrule
Llama-3-8B-Instruct & 0.5027 & 0.1810 & \underline{0.5098} & 0.4654 & 0.1232 & \textbf{0.4937} \\
Qwen2.5-7B-Instruct & 0.6182 & \textbf{0.2836} & 0.4182 & \textbf{0.6444} & \textbf{0.2515} & 0.4231 \\
\midrule

% --- Diffusion Language Models ---
\multicolumn{7}{c}{\textit{Diffusion Language Models}} \\
\midrule

LLaDA-8B-Base     & 0.3212 & 0.1323 & 0.4063 & 0.3891 & 0.1996 & 0.4300 \\
LLaDA-8B-Instruct & 0.4000 & 0.1944 & 0.4300 & 0.4363 & 0.1768 & 0.4489 \\
LLaDA-8B-SFT & 0.4510 & 0.1666 & 0.4425 & 0.4414 & 0.1694 & \underline{0.4737} \\
\rowcolor[HTML]{E8E8E8} 
LLaDA-8B-SemMask (Ours) & \underline{0.6438} & \underline{0.2628} & 0.4434 & 0.4521 & 0.2303 & 0.4500 \\
\midrule
Dream-7B-Base     & 0.4963 & 0.1172 & 0.4517 & 0.4632 & 0.1651 & 0.4184 \\
Dream-7B-Instruct & 0.5366 & 0.2137 & 0.4290 & 0.5141 & 0.1899 & 0.4252 \\
Dream-7B-SFT & 0.6268 & 0.1508 & 0.4982 & 0.5440 & 0.1503 & 0.4664 \\
\rowcolor[HTML]{E8E8E8} 
Dream-7B-SemMask (Ours) & \textbf{0.7085} & 0.2499 & \textbf{0.5178} & \underline{0.5568} & \underline{0.2402} & 0.4547 \\
\bottomrule
\end{tabular}
}
\caption{Main results of code generation evaluation on HumanEval and MBPP. The best results in each category are highlighted in \textbf{bold}, and the second-best results are marked with \underline{underlines}.}
\label{tab:code_gen}

\end{table*}

%% ---- Experiments ---- %%
\section{Experiments}

\subsection{Experimental Setups}
% prompt和过滤规则放在附录
\paragraph{Datasets.} We constructed two high-quality training datasets tailored for text writing and code generation tasks. 
For the text writing task, we sourced data from WritingPrompts~\cite{fan2018hierarchical} and applied GPT-5.1-based quality filtering, ultimately obtaining a subset of 10,000 samples.
For code generation task, we applied the same filtering strategy to OpenCodeInstruct~\cite{ahmad2025opencodeinstruct}, resulting in 10,000 high-quality training samples.
% Detailed data processing pipelines and filtering prompts are provided in Appendix~\ref{app:data_setup}.

% 放在附录
% \textbf{Implementation Details.} We employed LLaDA-8B-Instruct and Dream-7B-Instruct as backbones, fine-tuned via LoRA on all linear modules with a rank of 64 and an alpha of 128. Training was conducted for 3 epochs using the AdamW optimizer, utilizing a cosine scheduler peaking at $1 \times 10^{-4}$ and a global batch size of 32. The context window was fixed at 4096 tokens for both domains.GOD n=30

\paragraph{Baselines.} We benchmarked our approach against Llama-3-8B-Instruct~\cite{grattafiori2024llama} and Qwen2.5-7B-Instruct~\cite{qwen2_5} as state-of-the-art autoregressive references. 
For diffusion baselines, we evaluated LLaDA-8B-Base, LLaDA-8B-Instruct, Dream-7B-Base, and Dream-7B-Instruct. 
We also evaluated two model variants, LLaDA-8B-SFT and Dream-7B-SFT, obtained by directly fine-tuning their respective instruction-tuned models on our curated datasets using LoRA~\cite{hu2022lora}.

% 数据集介绍和prompt放在附录
\paragraph{Benchmarks and Metrics.} We evaluate model performance on both tasks using multiple evaluation metrics.
% For the text writing task, we evaluate on HANNA~\cite{chhun2022human} for narrative continuation and CommonGen~\cite{lin2020commongen} for keyword-constrained generation.
For the text writing task, we evaluate the model on two benchmarks: HANNA~\cite{chhun2022human}, which focuses on narrative continuation, and CommonGen~\cite{lin2020commongen}, which focuses on keyword-constrained generation.
We used BLEU and ROUGE-L to measure lexical overlap and surface similarity, BERT-Score and BLEURT to assess semantic consistency, and perplexity and GPT-5.1-Score to evaluate overall linguistic quality.
Additionally, we introduce the GOD metric~\cite{gwak2025reward} to evaluate the model's non-autoregressive generation behavior.
For the code generation task, we used HumanEval~\cite{chen2021evaluating} and MBPP~\cite{austin2021program} to evaluate functional correctness using accuracy, with CodeBLEU and BLEURT assessing structural and semantic similarity.
% Detailed descriptions of these benchmarks are available in Appendix. 

% ------ Results ------- %
\subsection{Main Results}
\paragraph{Text Writing.} As illustrated in Table~\ref{tab:writing_gen}, SemMask consistently improves semantic coherence across both LLaDA-based and Dream-based architectures.
On HANNA, Dream-7B-SemMask achieves the strongest overall performance with 0.4613 BLEU and 0.8254 BLEURT, outperforming the autoregressive baselines on most metrics, while Llama-3-8B-Instruct retains the highest BERT-Score of 0.8051. 
LLaDA-8B-SemMask also substantially improves over LLaDA-8B-SFT in both BLEU and GPT-5.1-Score.
On CommonGen, SemMask obtains markedly higher GOD scores and lower perplexity, with Dream-7B-SemMask achieving the best GOD score of 7.4544 and the lowest perplexity of 21.6630. 
These results indicate that SemMask restores the parallel generation capability of diffusion language models and generates more coherent text.

\paragraph{Code Generation.} We further evaluate SemMask on code generation to assess its generalization in highly structured domains. 
As shown in Table~\ref{tab:code_gen}, Dream-7B-SemMask achieves the best HumanEval accuracy of 0.7085, outperforming Dream-7B-Instruct at 0.5366 and the autoregressive Qwen2.5-7B-Instruct at 0.6182. 
LLaDA-8B-SemMask likewise improves HumanEval accuracy from 0.4000 to 0.6438 and surpasses Llama-3-8B-Instruct in CodeBLEU, indicating stronger syntactic and structural consistency. 
Although Qwen2.5-7B-Instruct remains the strongest model on MBPP accuracy, LLaDA-8B-SemMask raises MBPP CodeBLEU from 0.1768 to 0.2303, while Dream-7B-SemMask also improves BLEURT. 
These results demonstrate that SemMask enhances both functional correctness and structural integrity by better aligning natural-language instructions with executable code.

\subsection{Ablation Study}

\paragraph{Validation of Gradient-Based Semantic Token Scoring.} To validate gradient norms as a proxy for semantic importance, we compare SemMask with Random, IDF-based, POS-based, and Reverse Gradient strategies. 
For the POS baseline, we use spaCy to identify semantic tokens, including nouns, verbs, adjectives, and proper nouns, and prioritize their preservation. 
Reverse Gradient inverts our proposed selection rule by retaining tokens with the lowest gradient norms as semantic anchors.
As shown in Table~\ref{tab:ablation}, SemMask achieves the best semantic performance on both backbones, increasing BLEU over the POS baseline from 0.3641 to 0.3921 on LLaDA and from 0.4154 to 0.4613 on Dream.
Although Random achieves a marginally higher GOD score, its substantially weaker semantic metrics indicate that generation-order deviation alone does not guarantee coherent generation.

\begin{table}[t]

\centering
\resizebox{0.48\textwidth}{!}{%
\begin{tabular}{lcccccc}
\toprule
\textbf{Method} & \textbf{GOD}$\uparrow$ & \textbf{BLEU}$\uparrow$ & \textbf{BLEURT}$\uparrow$ & \textbf{ROUGE-L}$\uparrow$ & \textbf{BERT-S}$\uparrow$ \\
\midrule
\multicolumn{6}{c}{\textit{LLaDA-8B-Instruct}} \\
\midrule
Random & \textbf{1002.4} & 0.3667 & 0.7016 & 0.2194 & 0.7092 \\
IDF & 994.2 & 0.3532 & 0.7054 & 0.2445 & 0.7612 \\
POS & 997.5 & 0.3641 & 0.7118 & 0.2616 & 0.7659 \\
Reverse Gradient & 985.3 & 0.2105 & 0.5841 & 0.1552 & 0.5845 \\
% \rowcolor[HTML]{E6F0FF} 
\textbf{SemMask (Ours)} & 998.9 & \textbf{0.3921} & \textbf{0.7301} & \textbf{0.3101} & \textbf{0.7746} \\
\midrule
\multicolumn{6}{c}{\textit{Dream-7B-Instruct}} \\
\midrule
Random & \textbf{1005.1} & 0.3940 & 0.6972 & 0.3315 & 0.7664 \\
IDF & 996.5 & 0.3915 & 0.7754 & 0.3549 & 0.7690 \\
POS & 998.1 & 0.4154 & 0.7922 & 0.3754 & 0.7710 \\
Reverse Gradient & 988.4 & 0.2510 & 0.6555 & 0.2245 & 0.7085 \\
% \rowcolor[HTML]{E6F0FF} 
\textbf{SemMask (Ours)} & 999.8 & \textbf{0.4613} & \textbf{0.8254} & \textbf{0.4181} & \textbf{0.7762} \\
\bottomrule
\end{tabular}%
}
\caption{Ablation study of semantic token scoring strategies on HANNA. Both the generation length and the number of sampling steps are set to 3,000. The best result for each metric under each setting is highlighted in \textbf{bold}.}
\label{tab:ablation}
\end{table}

\begin{table}[t]
\centering
\resizebox{0.48\textwidth}{!}{%
\begin{tabular}{llcc}
\toprule
\textbf{Method} & \textbf{Strategy} & \textbf{GOD} $\uparrow$ & \textbf{Perplexity} $\downarrow$ \\
\midrule
\multicolumn{4}{c}{\textit{LLaDA-8B-Instruct}} \\
\midrule
Unconstrained Strategy & $\delta = 0$ & 3.8210 & 36.4215 \\
Static Constraint Strategy & $\delta = 2$ & 3.4520 & 26.1240 \\
Static Constraint Strategy & $\delta = 4$ & 4.9564 & 27.5412 \\
\textbf{SemMask} & $\delta_t\leftarrow\left\lfloor\delta t\right\rfloor$ & \textbf{6.3187} & \textbf{22.5286} \\
\midrule
\multicolumn{4}{c}{\textit{Dream-7B-Instruct}} \\
\midrule
Unconstrained Strategy & $\delta = 0$ & 2.9004 & 34.1200 \\
Static Constraint Strategy & $\delta = 2$ & 5.1500 & 25.1801 \\
Static Constraint Strategy & $\delta = 4$ & 5.2810 & 26.9499 \\
\textbf{SemMask} & $\delta_t\leftarrow\left\lfloor\delta t\right\rfloor$ & \textbf{7.4544} & \textbf{21.6630} \\
\bottomrule
\end{tabular}%
}
\caption{Ablation study of anchor distance constraint strategies on CommonGen. Both the generation length and the number of sampling steps are set to 30. The best result for each metric under each setting is highlighted in \textbf{bold}.}
\label{tab:ablation_masking}
\end{table}

\paragraph{Effectiveness of Semantic Anchor Preservation Masking.} To evaluate Semantic Anchor Preservation Masking, we compare SemMask with an unconstrained strategy and a static constraint strategy using two fixed anchor distances on CommonGen.
The unconstrained strategy yields the weakest overall performance, suggesting that clustered semantic anchors provide insufficient contextual coverage. 
The two static constraint strategies improve performance, but their fixed anchor distances remain suboptimal across different masking ratios and sequence structures. 
As shown in Table~\ref{tab:ablation_masking}, SemMask achieves the highest GOD and lowest perplexity on both backbones, reaching 6.3187 and 22.5286 on LLaDA and 7.4544 and 21.6630 on Dream, respectively. 
These results indicate that adapting the anchor distance to the masking ratio better balances non-autoregressive generation behavior and linguistic fluency.

% \begin{figure}[t]
%     \centering
%     \includegraphics[width=\linewidth]{imgs/step3.pdf}
%     \caption{Performance of LLaDA-based and Dream-based models on HumanEval and MBPP benchmarks across varying diffusion steps.}
%     % \vspace{-2mm}
%     \label{fig:step_analysis}
% \end{figure}

% 附录
% The semantic topology is computed using the frozen pre-trained weights of each respective backbone."

% GOD
% Standard measure of text fluency
% and coherence, particularly relevant in keywordconstrained tasks.
% keyword-constrained generation task designed to
% test non-autoregressive capabilities.
% \subsection{参数的分析}

\begin{figure*}[h]
    \centering
    \includegraphics[width=\linewidth]{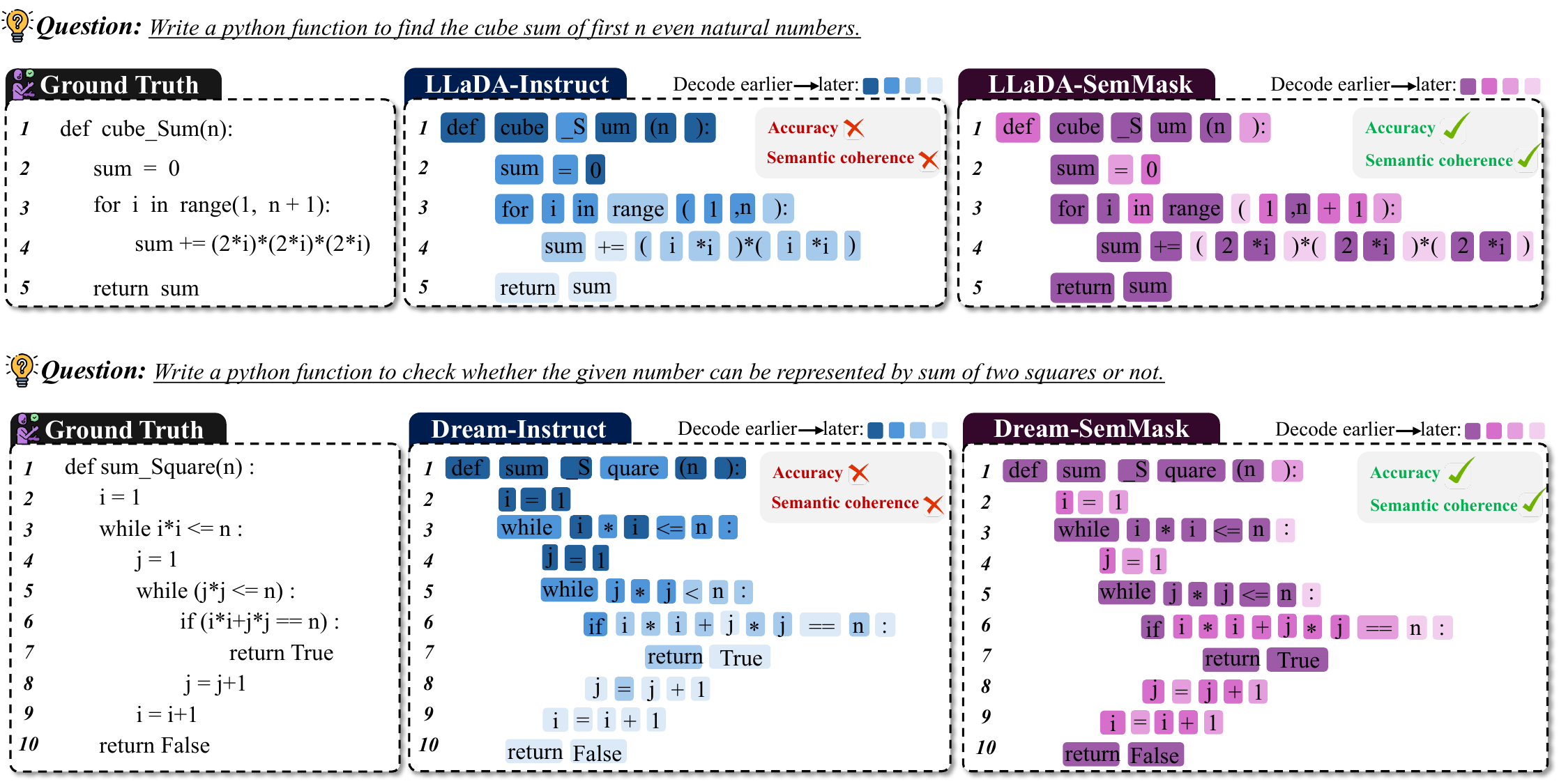}
    \caption{Case study on MBPP comparing the generated programs and token decoding orders of the Instruct baselines and SemMask.}
    % \vspace{-1mm}
    \label{fig:case}
\end{figure*}

% \begin{figure}[t]
%     \centering
%     \includegraphics[width=0.94\linewidth]{Figures/per.pdf}
% \caption{Efficiency--accuracy trade-off on HumanEval with a sequence length of 512 on $4\times$V100 GPUs. Bubble size denotes the average number of decoding steps.}
% \label{fig:efficiency_accuracy}
% \end{figure}

\begin{figure}[t]
    \centering
    \includegraphics[width=0.80\linewidth]{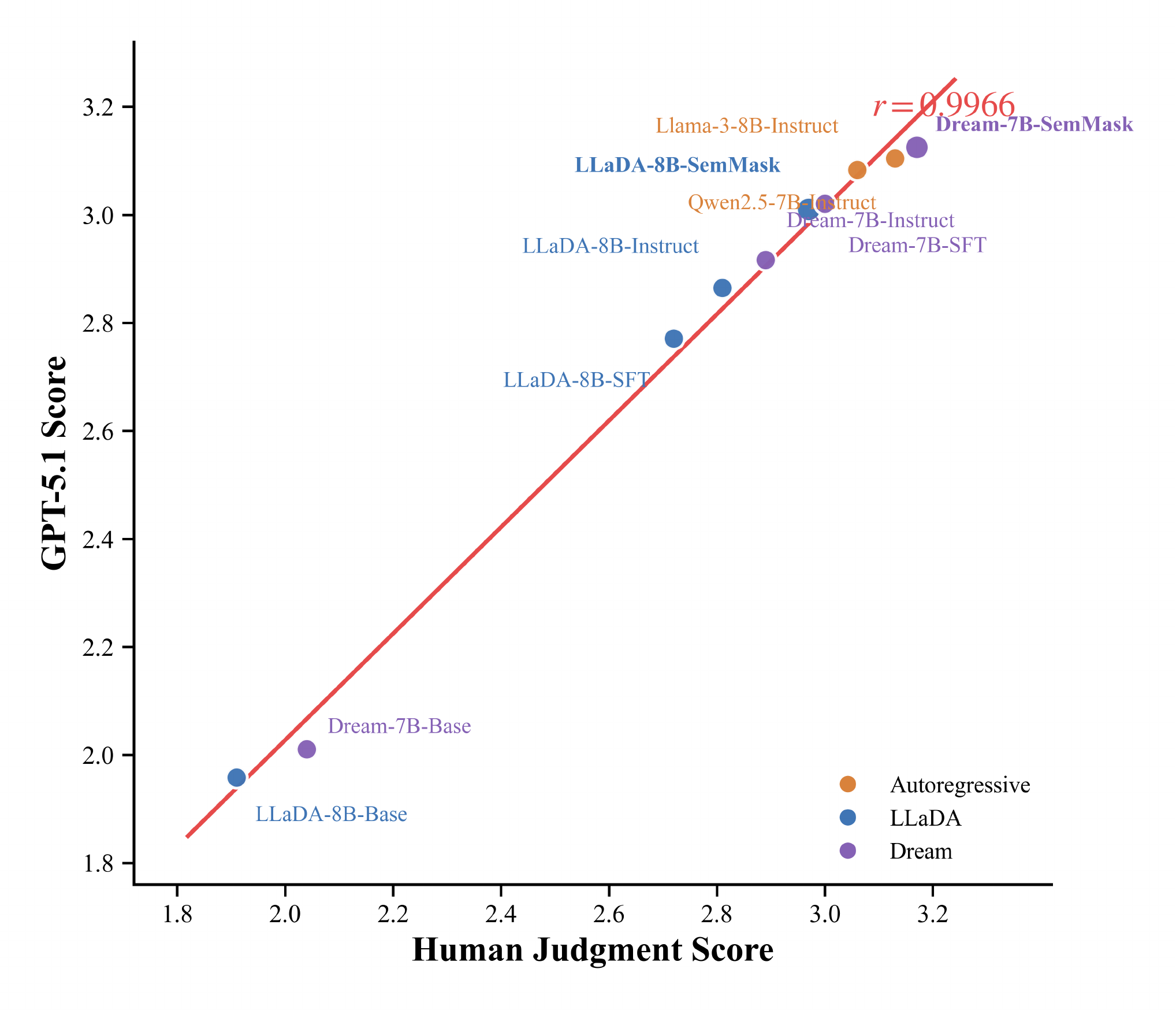}
    \caption{ Automatic evaluation score vs. human rating.}
    \vspace{-4mm}
    % \vspace{-2mm}
    \label{fig:human}
\end{figure}

\begin{table}[t]
\centering
\resizebox{0.42\textwidth}{!}{%
\begin{tabular}{lccc}
\toprule
\textbf{Method}
& \textbf{\#Steps} $\downarrow$
& \textbf{\#Time} $\downarrow$
& \textbf{Acc} $\uparrow$ \\
\midrule

\multicolumn{4}{c}{\textit{LLaDA-8B}} \\
\midrule
Base                  & 441          & 23.54s         & 0.3212          \\
Instruct              & 323          & 16.49s         & 0.4000          \\
SFT                   & 214          & 11.12s         & 0.4510          \\
\textbf{SemMask (Ours)}
                      & \textbf{166} & \textbf{8.73s} & \textbf{0.6438} \\
\midrule

\multicolumn{4}{c}{\textit{Dream-7B}} \\
\midrule
Base                  & 439          & 20.08s         & 0.4963          \\
Instruct              & 340          & 15.52s         & 0.5366          \\
SFT                   & 201          & 9.51s          & 0.6268          \\
\textbf{SemMask (Ours)}
                      & \textbf{169} & \textbf{7.87s} & \textbf{0.7085} \\
\bottomrule
\end{tabular}%
% \vspace{-2mm}
}
% \caption{Comparison of generation efficiency and accuracy. The best result for each metric within each model family is highlighted in \textbf{bold}.}
\caption{Comparison of generation efficiency and accuracy. SemMask substantially reduces both the average number of decoding steps and inference time while achieving the highest accuracy. The best result for each metric within each model family is highlighted in \textbf{bold}.}
\label{tab:efficiency_accuracy}
\end{table}

\subsection{Efficiency--Accuracy Trade-off Analysis}

% To assess inference efficiency, we report the per-sample runtime and average decoding steps of each method under Appendix B.2 configuration on HumanEval with a sequence length of 512, measured on $4\times$V100 GPUs. 
To assess inference efficiency, we report the per-sample runtime and average number of decoding steps for each method on HumanEval, using the configuration described in Appendix B.2 and a sequence length of 512. All runtimes are measured on $4\times$ V100 GPUs.
As shown in Table~\ref{tab:efficiency_accuracy}, SemMask consistently achieves higher accuracy with fewer decoding steps and shorter inference time across both backbones.
Compared with LLaDA-8B-SFT, LLaDA-8B-SemMask reduces the average decoding steps from 214 to 166 and the inference time from 11.12\,s to 8.73\,s while improving accuracy from 0.4510 to 0.6438; Dream-7B-SemMask similarly achieves an accuracy of 0.7085 using only 169 steps and 7.87\,s. 
% These results demonstrate that SemMask simultaneously improves generation efficiency and accuracy.

\subsection{Case Study and Consistency with Human Judgments}

Figure~\ref{fig:case} presents two representative MBPP cases, where our method, SemMask, preserves critical arithmetic operations and boundary conditions that are incorrectly generated by the Instruct baselines. 
The visualized decoding orders further indicate that SemMask prioritizes semantically important tokens and establishes coherent program structures earlier, thereby improving functional correctness and semantic coherence.
Figure~\ref{fig:human} shows a strong correlation between GPT-5.1 scores and human judgments ($r=0.9966$), confirming the reliability of the automatic evaluation and the superior performance of SemMask.

\section{Conclusion}

In this work, we introduce SemMask, a training paradigm designed to enhance the parallel generation capabilities of DLMs by mitigating pseudo-autoregressive behavior. 
Its Gradient-Based Semantic Token Scoring and Semantic Anchor Preservation Masking encourage important semantic tokens to be generated earlier, establishing a global semantic scaffold for coherent generation. 
Experiments on text and code generation tasks demonstrate consistent improvements in semantic coherence, generation efficiency, and task performance over standard DLM baselines. 
These results establish SemMask as an effective and robust approach for scalable non-autoregressive sequence generation.

\bibliography{aaai2027}

\newpage

\section*{Appendix Overview}
This appendix includes the following sections:
\begin{itemize}
    \item Section A introduces the preliminaries of masked diffusion language models, including the forward masking process, training objective, and inference process.
    \item Section B provides additional implementation details, including data processing and filtering, training and inference settings, gradient score computation, and the human evaluation protocol for HANNA.
    \item Section C presents additional experimental results, including computational overhead analysis, the adaptation of SemMask to continuous diffusion models, and generalization to mathematical reasoning tasks.
    \item Section D provides additional case studies illustrating how SemMask prioritizes semantically important tokens and establishes the core semantic structure early in the decoding process.
\end{itemize}

\subsection*{A. Preliminaries}

\paragraph{Masked Diffusion Language Models.}
Unlike autoregressive language models that generate tokens in a strict left-to-right order, masked diffusion language models formulate generation as a probabilistic process consisting of a forward masking corruption process and a reverse denoising recovery process.
Given a clean sequence $\boldsymbol{x}_0=(x_0^1,\ldots,x_0^L)$ and a masking ratio $t$, the forward process independently replaces each token with \texttt{[M]} with probability $t$:
\begin{equation}
q(\boldsymbol{x}_t \mid \boldsymbol{x}_0)
=
\prod_{i=1}^{L}
\left[
(1-t)\delta(x_t^i=x_0^i)
+
t\delta(x_t^i=\texttt{[M]})
\right]
\end{equation}
The reverse process is parameterized by a mask predictor $p_\theta$, which takes the corrupted sequence $\boldsymbol{x}_t$ as input and jointly predicts the original tokens at all masked positions:
\begin{equation}
p_\theta(\boldsymbol{x}_0 \mid \boldsymbol{x}_t)
=
\prod_{i:x_t^i=\texttt{[M]}}
p_\theta(x_0^i \mid \boldsymbol{x}_t)
\end{equation}
The model is optimized using a weighted cross entropy loss restricted to the masked positions:
\begin{equation}
\resizebox{1\columnwidth}{!}{$
\mathcal{L}(\theta)
=
-\mathbb{E}_{t,\boldsymbol{x}_0,\boldsymbol{x}_t}
\left[
\frac{1}{t}
\sum_{i=1}^{L}
\mathbb{I}[x_t^i=\texttt{[M]}]
\log p_\theta(x_0^i \mid \texttt{prompt}, \boldsymbol{x}_t)
\right]
$}
\end{equation}
where $t\sim\mathcal{U}(0,1)$, $\boldsymbol{x}_0$ is sampled from the training data, and $\boldsymbol{x}_t$ is sampled from the forward masking process.
During supervised fine tuning, the prompt tokens remain unchanged, while only the response tokens are independently masked and included in the training objective.

\paragraph{Inference Process.}
The inference process consists of a sequence of denoising steps.
At each step, the model predicts all currently masked tokens in parallel, keeps reliable predictions, and remasks uncertain positions for further refinement.
This parallel update mechanism allows multiple tokens to be generated simultaneously, offering greater flexibility and potentially higher efficiency than autoregressive decoding.

\subsection*{B. More Implementation Details}
% Add content here...
\subsubsection*{B.1. Data Processing and Filtering}

\paragraph{Text Writing Training Data.} To construct a high quality training dataset for text writing, we collected pairs of prompts and stories from the WritingPrompts dataset. 
To prevent potential data leakage, we used scripts based on predefined rules to remove samples overlapping with the HANNA benchmark. 
We then applied heuristic filtering to eliminate invalid or mixed-language text, nonstandard special characters, and samples exceeding the length limit of 4,096 tokens. 
We further employed GPT-5.1 to evaluate grammatical fluency, narrative coherence, logical consistency, content completeness, and alignment between each prompt and its corresponding story, removing samples with awkward expressions, abrupt plot transitions, contradictory events, or incomplete narratives. 
The resulting dataset contains 10,000 high-quality samples for training for the text writing task.

\paragraph{Code Generation Training Data.} For the code generation task, we collected code solution pairs from the OpenCodeInstruct dataset. 
We removed samples overlapping with the HumanEval and MBPP benchmarks to prevent potential data leakage. 
We also excluded samples exceeding the length limit of 4,096 tokens and employed GPT-5.1 to evaluate the logical correctness and functional consistency of each code solution. 
This processing pipeline resulted in a training dataset containing 10,000 code solution pairs.

\subsubsection*{B.2. Training and Inference Settings}
\label{app:setting}

\begin{table*}[t]
\centering
% \caption{Training configurations for SemMask across the two backbone models.}
% \label{tab:training_config}
\resizebox{\linewidth}{!}{%
\begin{tabular}{l|cccccc}
\toprule
\textbf{Base Model} & \textbf{LoRA Rank} & \textbf{LoRA Alpha} &
\textbf{Learning Rate} & \textbf{LR Schedule} &
\textbf{Batch Size} & \textbf{Epochs} \\
\midrule
LLaDA-8B-Instruct & 64 & 64 & $2\times10^{-4}$ & cosine & 32 & 3 \\
Dream-7B-Instruct & 32 & 32 & $2\times10^{-4}$ & cosine & 32 & 3 \\
\bottomrule
\end{tabular}%
}
\caption{Training configurations for SemMask across the two backbone models.}
\label{tab:training_config}
\end{table*}

\begin{table*}[t]
\centering
\resizebox{\linewidth}{!}{%
\begin{tabular}{l|lccccc}
\toprule
\textbf{Model} & \textbf{Phase} & \textbf{Total Time (s)} &
\textbf{Time per Step (s)} & \textbf{Total PFLOPs} &
\textbf{Throughput (TFLOPs/s)} \\
\midrule
LLaDA-8B-SFT
& Training
& 21,502.79
& 50.00
& 158.42
& 7.37 \\

LLaDA-8B-SemMask (Ours)
& Preprocessing
& 392.56
& --
& 34.21
& 87.15 \\

LLaDA-8B-SemMask (Ours)
& Training
& 21,620.88
& 50.00
& 159.90
& 7.40 \\
\midrule
Dream-7B-SFT
& Training
& 27,180.11
& 62.50
& 138.62
& 5.10 \\

Dream-7B-SemMask (Ours)
& Preprocessing
& 428.33
& --
& 36.29
& 84.72 \\

Dream-7B-SemMask (Ours)
& Training
& 27,399.28
& 66.66
& 138.62
& 5.06 \\
\bottomrule
\end{tabular}%
}
\caption{Comparison of preprocessing and training costs between standard SFT and SemMask on 10,000 code training samples using $4\times$V100 GPUs.}
\label{tab:computational_overhead}
\end{table*}

As shown in Table~\ref{tab:training_config}, we present the training configurations for both backbone models, with a context length of 4,096 tokens and LoRA applied to all linear projection layers.
The semantic importance scores are computed using the frozen weights of each backbone, and SemMask uses $\lambda=0.1$, $\delta=4$, $\alpha=10.0$, and $\beta=2.0$.
During inference, we adopt pure diffusion sampling with low-confidence remasking for LLaDA and a maximum threshold of 0.6 for Dream, setting both the generation length and sampling steps to 30 for CommonGen, 3,000 for HANNA, and 512 for HumanEval and MBPP.

\subsubsection*{B.3. Gradient Score Computation}

For each sample, we sample multiple masking ratios $t\sim\mathcal{U}(0,1)$ and independently mask the response tokens with probability $t$, while keeping the prompt unchanged.
Using the frozen backbone, we compute the original denoising objective only over the masked response tokens in each corrupted version.
For every visible response token, we calculate the gradient of the denoising loss with respect to its input embedding and project this gradient onto the direction from the original token embedding to the mask token embedding.
This projection measures how strongly masking the visible token would affect the reconstruction loss over the other masked tokens; a larger value indicates a greater contextual contribution and thus a higher semantic importance score.
We average the scores across multiple masked versions to reduce randomness, such that tokens contributing more to the reconstruction of the masked sequence receive higher scores.

\subsubsection*{B.4. Human Evaluation Details}

To evaluate text generation quality on the HANNA benchmark, three evaluators with master's degrees independently rated each generated text on a five point scale. The evaluation considers five dimensions: content completeness, relevance, coherence, fluency, and factual consistency. The scoring criteria are defined as follows:

\begin{itemize}
    \item \textbf{1:} The text fails to address the prompt or is largely incomprehensible, with severe deficiencies in at least four evaluation dimensions.
    \item \textbf{2:} The text partially addresses the prompt but contains major problems in at least three dimensions, such as substantial content omissions, logical contradictions, frequent grammatical errors, or factual inconsistencies.
    \item \textbf{3:} The text addresses the main topic and remains understandable, but contains one or two major deficiencies or several noticeable problems in completeness, coherence, fluency, relevance, or factual consistency.
    \item \textbf{4:} The text fully addresses the prompt and contains no major deficiencies, with only minor issues in one or two evaluation dimensions that do not substantially affect overall quality.
    \item \textbf{5:} The text fully satisfies the prompt across all five evaluation dimensions, demonstrating complete content, strong relevance, coherent organization, fluent language, and factual consistency without notable errors.
\end{itemize}

For each test item $i$ and evaluation dimension $d$, the mean score assigned by the three evaluators is calculated as:
\begin{equation}
\bar{S}_{i}^{(d)}
=
\frac{S_{i,1}^{(d)}+S_{i,2}^{(d)}+S_{i,3}^{(d)}}{3}
\end{equation}
where $S_{i,1}^{(d)}$, $S_{i,2}^{(d)}$, and $S_{i,3}^{(d)}$ denote the scores assigned by the three evaluators.

The overall score for item $i$ is computed by averaging across the five evaluation dimensions:
\begin{equation}
\bar{S}_{i}
=
\frac{1}{5}
\sum_{d=1}^{5}\bar{S}_{i}^{(d)}
\end{equation}

The final human evaluation score for a model over all $N$ HANNA test samples is:
\begin{equation}
S_{\mathrm{final}}
=
\frac{1}{N}
\sum_{i=1}^{N}\bar{S}_{i}
\end{equation}

\subsection*{C. More Experimental Results}
\subsubsection*{C.1. Computational Overhead Analysis}

To evaluate the additional computational cost of SemMask, we compare its preprocessing and training overhead with standard SFT on 10,000 code training samples using $4\times$V100 GPUs.
As shown in Table~\ref{tab:computational_overhead}, computing semantic importance scores requires only 392.56,s for LLaDA and 428.33,s for Dream.
During training, SemMask increases the total runtime by only 0.55\% for LLaDA and 0.81\% for Dream compared with standard SFT, while maintaining comparable computational throughput.
These results demonstrate that the additional cost of SemMask is mainly confined to the one-time preprocessing stage, with only marginal overhead during model training.

\subsubsection*{C.2. Adapting SemMask to Continuous Diffusion Models}

\begin{table}[t]
\centering
\resizebox{\linewidth}{!}{%
\begin{tabular}{l|ccccc}
\toprule
\textbf{Model} & \textbf{BLEU} & \textbf{BLEURT} &
\textbf{ROUGE-L} & \textbf{BERT-S} &
\textbf{GPT-5.1-S} \\
\midrule
DiffuSeq-SFT
& 0.2546
& 0.4368
& 0.1001
& 0.5389
& 1.2658 \\

DiffuSeq-SemMask (Ours)
& \textbf{0.2912}
& \textbf{0.4490}
& \textbf{0.1424}
& \textbf{0.5918}
& \textbf{1.9644} \\
\midrule
DiffuSeq-v2-SFT
& 0.2792
& 0.4567
& 0.1968
& 0.5674
& 1.5349 \\

DiffuSeq-v2-SemMask (Ours)
& \textbf{0.3106}
& \textbf{0.5004}
& \textbf{0.2165}
& \textbf{0.5977}
& \textbf{2.0116} \\
\bottomrule
\end{tabular}%
}
\caption{Performance of SemMask adapted to continuous diffusion models on the HANNA benchmark.}
\label{tab:continuous_hanna}
\end{table}

\begin{table}[t]
\centering
\resizebox{\columnwidth}{!}{%
\begin{tabular}{l|ccc}
\toprule
\textbf{Model} & \textbf{Accuracy} &
\textbf{CodeBLEU} & \textbf{BLEURT} \\
\midrule
DiffuSeq-SFT
& 0.1644
& 0.0921
& 0.2397 \\

DiffuSeq-SemMask (Ours)
& \textbf{0.1947}
& \textbf{0.1166}
& \textbf{0.2963} \\
\midrule
DiffuSeq-v2-SFT
& 0.2095
& 0.0990
& 0.2978 \\

DiffuSeq-v2-SemMask (Ours)
& \textbf{0.2549}
& \textbf{0.1369}
& \textbf{0.3259} \\
\bottomrule
\end{tabular}%
}
\caption{Performance of SemMask adapted to continuous diffusion models on HumanEval.}
\label{tab:continuous_humaneval}
\end{table}

\begin{figure*}[ht]
    \centering
    \includegraphics[width=\linewidth]{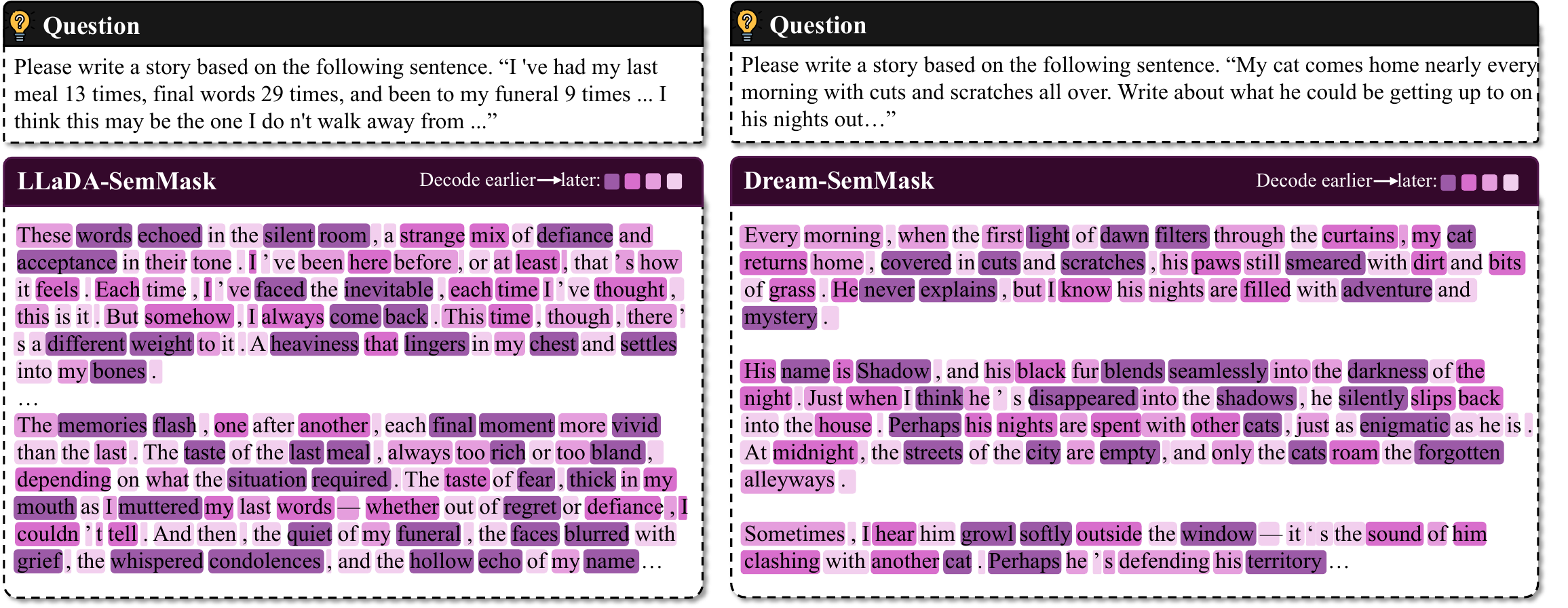}
    \caption{Case studies of SemMask on the HANNA benchmark.}
    % \vspace{-1mm}
    \label{fig:case3}
\end{figure*}

\begin{figure*}[ht]
    \centering
    \includegraphics[width=\linewidth]{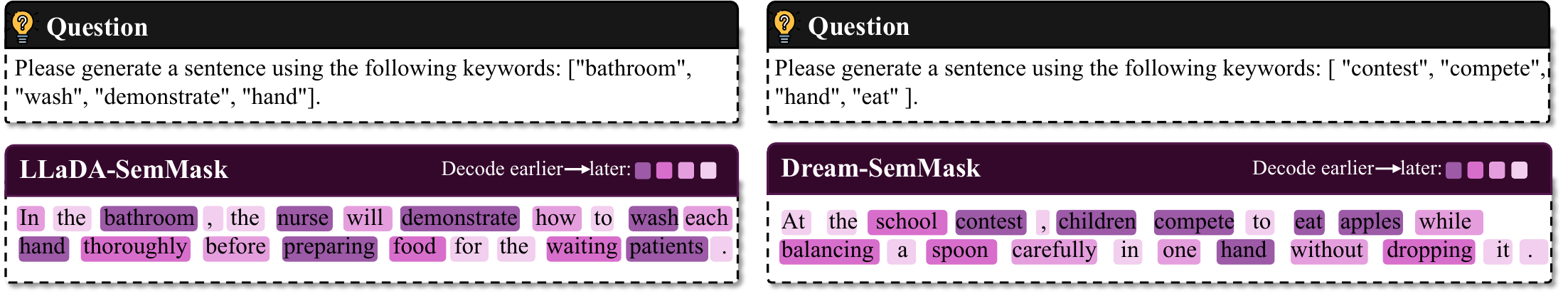}
    \caption{Case studies of SemMask on the CommonGen benchmark.}
    \vspace{-1mm}
    \label{fig:case2}
\end{figure*}

We adapt SemMask to two representative continuous diffusion models, DiffuSeq and DiffuSeq-v2, by applying semantic guidance only to the target sequence.
During training, high-importance target tokens receive weaker corruption and larger denoising weights, while the original model architectures and sampling mechanisms are otherwise preserved.
During inference, high-scoring target tokens are fixed earlier to establish a global semantic structure before function words and local details are progressively completed.
As shown in Tables~\ref{tab:continuous_hanna} and~\ref{tab:continuous_humaneval}, SemMask consistently improves all evaluation metrics on both HANNA and HumanEval, with DiffuSeq-v2-SemMask achieving 0.3106 BLEU and 0.2549 accuracy.
These results demonstrate that SemMask can be effectively extended beyond masked discrete diffusion models to continuous diffusion frameworks.

\subsubsection*{C.3. Generalization to Mathematical Reasoning Tasks}

\begin{table}[t]
\centering
\resizebox{\columnwidth}{!}{%
\begin{tabular}{l|cc}
\toprule
\textbf{Model} & \textbf{GSM8K} $\uparrow$ & \textbf{MATH-500} $\uparrow$ \\
\midrule
LLaDA-8B-Base
& 67.84
& 28.65 \\

LLaDA-8B-Instruct
& 68.21
& 30.47 \\

LLaDA-8B-SFT
& 70.09
& 31.93 \\

LLaDA-8B-SemMask (Ours)
& \textbf{73.20}
& \textbf{31.96} \\
\midrule
Dream-7B-Base
& 74.37
& 36.55 \\

Dream-7B-Instruct
& 75.08
& 37.10 \\

Dream-7B-SFT
& 76.89
& \textbf{39.66} \\

Dream-7B-SemMask (Ours)
& \textbf{78.28}
& 39.58 \\
\bottomrule
\end{tabular}%
}
\caption{Performance comparison on the GSM8K and MATH-500 mathematical reasoning benchmarks. The best result within each model family is highlighted in bold.}
\label{tab:math_generalization}
\end{table}

\begin{figure*}[t]
    \centering
    \includegraphics[width=\linewidth]{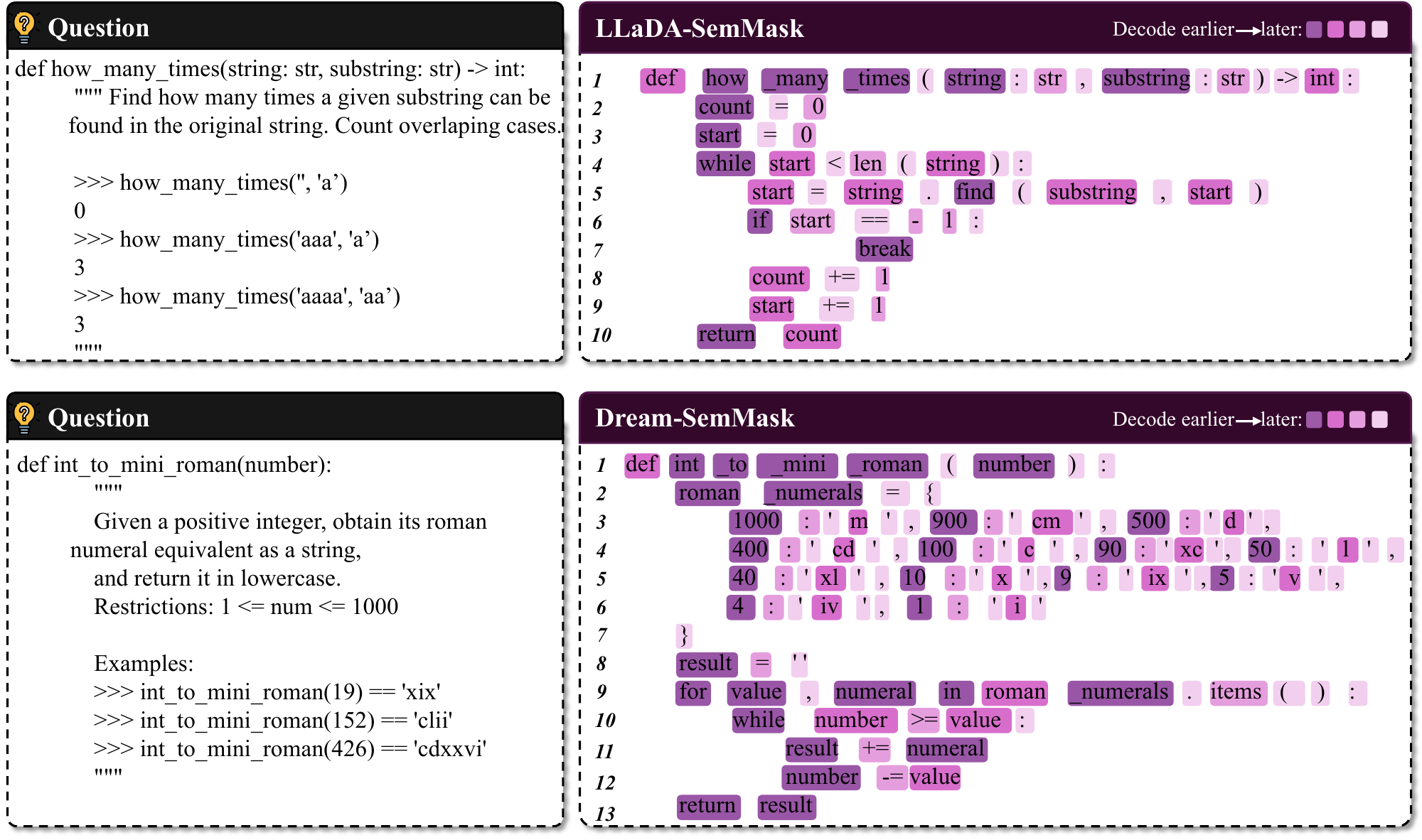}
    \caption{Case studies of SemMask on the HumanEval benchmark.}
    % \vspace{-1mm}
    \label{fig:case1}
\end{figure*}
\vspace{-2mm}

% \begin{figure*}[ht]
%     \centering
%     \includegraphics[width=\linewidth]{Figures/case14.pdf}
%     \caption{Case studies of SemMask on the CommonGen benchmark.}
%     \vspace{-1mm}
%     \label{fig:case2}
% \end{figure*}

% \begin{figure*}[ht]
%     \centering
%     \includegraphics[width=\linewidth]{Figures/case15.pdf}
%     \caption{Case studies of SemMask on the HANNA benchmark.}
%     % \vspace{-1mm}
%     \label{fig:case3}
% \end{figure*}

We further evaluate the generalization of SemMask to mathematical reasoning tasks using the GSM8K and MATH-500 benchmarks.
As shown in Table~\ref{tab:math_generalization}, SemMask improves GSM8K accuracy from 70.09 to 73.20 on LLaDA and from 76.89 to 78.28 on Dream.
On MATH-500, SemMask achieves performance comparable to the corresponding SFT baselines, with only marginal differences on both backbones.
These results suggest that semantic-prioritized generation is particularly beneficial for longer reasoning trajectories, while offering limited advantages for tasks with shorter outputs.

\subsection*{D. More Cases}

We present additional case studies in Figure~\ref{fig:case3}, Figure~\ref{fig:case2} and Figure~\ref{fig:case1}. 
These cases further demonstrate that SemMask prioritizes semantically important tokens, enabling the model to establish a global semantic scaffold early and progressively complete the remaining tokens.

\end{document}